\documentclass[lettersize,journal]{IEEEtran}
\usepackage{amsmath,amsfonts}
\usepackage{algorithmic}
\usepackage{algorithm}
\usepackage{array}
\usepackage[caption=false,font=normalsize,labelfont=sf,textfont=sf]{subfig}
\usepackage{textcomp}
\usepackage{stfloats}
\usepackage{url}
\usepackage{verbatim}
\usepackage{graphicx}
\usepackage{hyperref}
\usepackage{cite}
\usepackage{amsmath}
\usepackage{bm}

\graphicspath{{figures/}{pictures/}{images/}{./}} 
\usepackage{times}                        
\usepackage{tabu}                      
\usepackage{booktabs}                  
\usepackage{lipsum}                    
\usepackage{mwe}                       

\usepackage{pifont}
\usepackage{fontawesome5} 

\usepackage{mathptmx}                  
\usepackage{booktabs}     
\usepackage{multirow}     
\usepackage{caption}      
\usepackage{fdsymbol}
\usepackage{enumitem}

\usepackage{float} 
\usepackage{booktabs}       
\usepackage[table]{xcolor}  
\usepackage{multirow}       
\usepackage{array}          
\usepackage{graphicx}       
\usepackage{tikz}
\usetikzlibrary{
    positioning,    
    arrows.meta,    
    shapes.geometric, 
    calc,           
    decorations.pathmorphing 
}
\usepackage{color}
\usepackage{xcolor}
\newcommand{\benchmark}{{\textsc{VisDeception}}\xspace}
\definecolor{seagreen}{HTML}{2E8B85}

\begin{document}

\title{Chart Deception in Vision–Language Models: \\From Vulnerability to Mitigation}

\author{
Ridwan Mahbub,
Mohammed Saidul Islam, \\
Md Tahmid Rahman Laskar$^{*}$,
Mizanur Rahman$^{*}$,
Mir Tafseer Nayeem$^{*}$,
and Enamul Hoque%
\thanks{$^{*}$Equal contribution.}%
\thanks{Ridwan Mahbub, Mohammed Saidul Islam, Md Tahmid Rahman Laskar, Mizanur Rahman, and Enamul Hoque are with York University.

E-mail: \{rmahbub, saidulis, tahmid20, mizanurr, enamulh\}@yorku.ca}%
\thanks{Mir Tafseer Nayeem is with the University of Alberta.

E-mail: mnayeem@ualberta.ca}%
}

\markboth{Journal of \LaTeX\ Class Files,~Vol.~14, No.~8, August~2021}%
{Shell \MakeLowercase{\textit{et al.}}: A Sample Article Using IEEEtran.cls for IEEE Journals}


\maketitle

\begin{abstract}
Information visualizations are widely used to communicate patterns, trends, and outliers, yet deceptive design choices---such as truncated or inverted axes, distorted aspect ratios, inappropriate encodings, and misleading color mappings---can systematically alter interpretation while preserving the underlying data. As Vision–Language Models (VLMs) are increasingly used for chart understanding and analytical reasoning, assessing their robustness to such deceptive visualizations has become critical for trustworthy data analysis.
We introduce \textit{VisDeception}, the first controlled paired benchmark for evaluating the robustness of VLMs to misleading chart designs. The benchmark contains 1,600 paired faithful and misleading charts spanning eight major categories of deceptive visualization tactics, where each misleading chart is paired with a faithful counterpart generated from the same underlying data. To isolate deception-induced reasoning errors from baseline chart-understanding errors, we introduce the \emph{Deception Score}, a paired evaluation metric that quantifies how misleading visualizations shift model responses away from the faithful interpretation of the data. Across  32,000 responses from 10 state-of-the-art VLMs, we find that even advanced models remain highly vulnerable to deceptive visual manipulations, particularly those involving axis transformations, misleading encodings, and color semantics.   To improve robustness, we further propose an inference-time multi-agent mitigation framework that grounds reasoning in structured chart metadata extracted from the visualization before answer generation, enabling models to reduce the influence of deceptive visual cues without requiring explicit user instructions about potential deception. 
Across most evaluated models, this approach reduces deception-induced error, with the largest gains observed for stronger VLMs (e.g., Gemini 2.5 Pro reduces its Deception Score from 0.099 to 0.024, a 76\% reduction). Together, our findings reveal important reliability gaps in current chart-understanding systems and establish benchmark-driven evaluation, deception-aware metrics, and structured reasoning as promising directions for developing more trustworthy VLMs for visual analytics.
\end{abstract}

\begin{IEEEkeywords}
Misleading Visualizations, Large Language Models, Vision Language Models, Taxonomy, Evaluation
\end{IEEEkeywords}

\begin{figure}[t]
    \centering
    \includegraphics[width=\columnwidth]{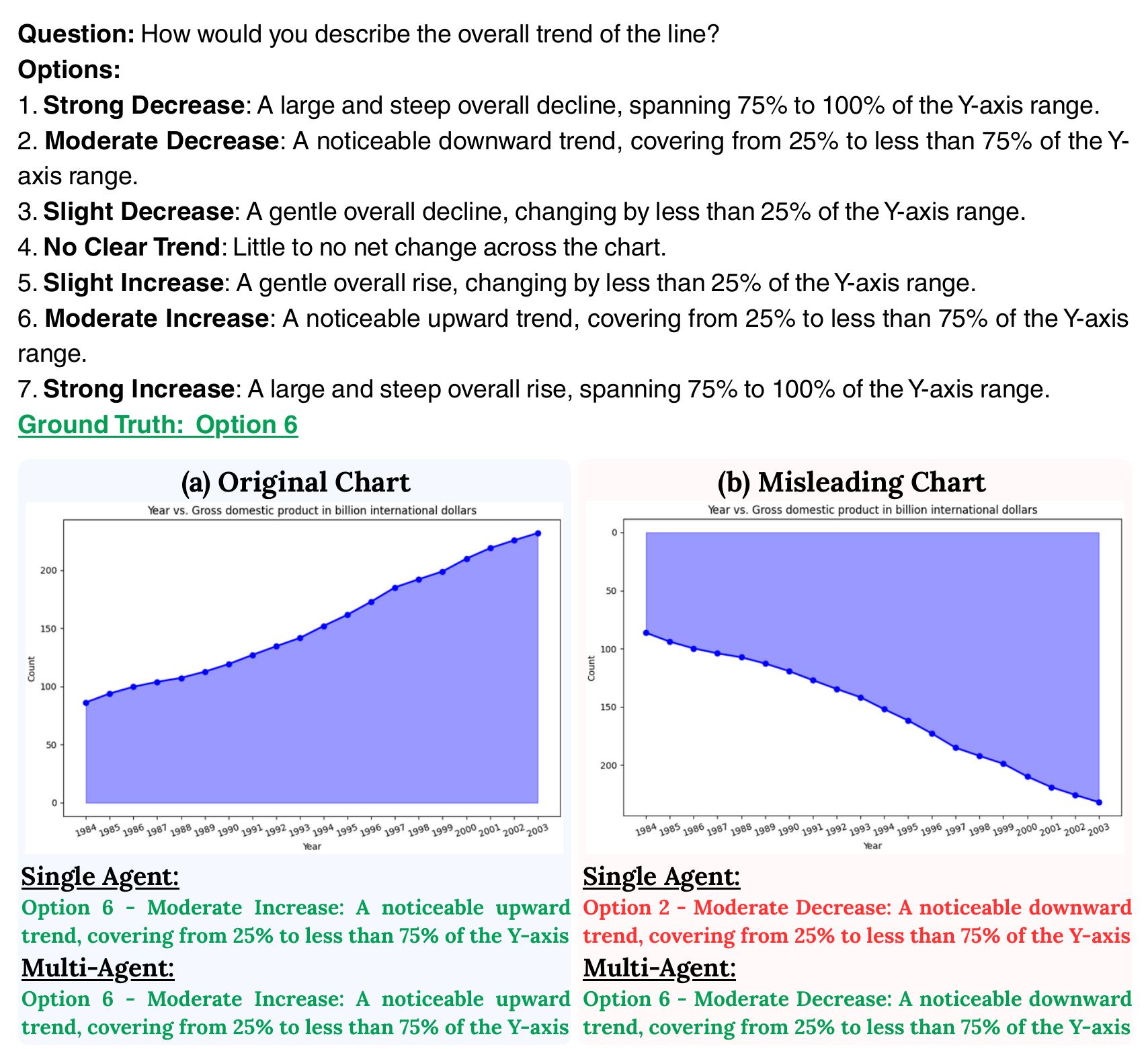}
    \caption{Misleading visual design can alter VLM reasoning without changing the underlying data. Both charts contain identical data and share the same ground-truth answer (\textbf{Option 6: Moderate Increase}). Yet the inverted-axis manipulation in (b) causes Gemini-2.5-Pro to incorrectly reverse the trend and predict \textbf{Option 2: Moderate Decrease}. Our structured-grounding mitigation framework restores the correct prediction for both charts, illustrating VLM susceptibility to misleading visualizations and the potential of inference-time mitigation.
    }
    \label{fig:teaser_mitigation}
\end{figure}
\section{Introduction}

Visualizations play a central role in how people interpret and communicate complex data by translating abstract numerical information into accessible visual narratives. By helping audiences identify patterns, trends, relationships, and anomalies, they have become essential to data-driven communication across domains such as journalism, public policy, healthcare, finance, and social media. As a result, visualizations not only support data storytelling but also shape interpretation and decision-making in high-stakes contexts \cite{segel2010narrative}. However, this communicative power can also be a double-edged sword. Misleading visual designs, such as truncated axes, skewed aspect ratios, and gratuitous three-dimensional embellishments, can distort perception without changing the underlying data \cite{huff2023lie, cairo2019charts, 10.1145/3613904.3642448}. Such designs are often intended to exaggerate or downplay differences between visual elements \cite{pandey2015deceptive}. Rather than fabricating facts, they manipulate how facts are visually represented, thereby amplifying or minimizing perceived differences and potentially shaping narratives and decisions in misleading ways \cite{huff2023lie, cairo2019charts, 10.1145/3613904.3642448}.

The risks become even more pronounced when visualizations are used within real-world decision-making systems. In domains such as finance, healthcare, policymaking, and enterprise reporting, AI-interpreted visualizations increasingly support tasks such as investment decisions, risk assessment, strategic planning, and public communication \cite{stadler2016improving, uddin2024data}. On social media, visualizations can also shape public discourse and influence collective understanding of complex issues \cite{chen2017social}. Prior work has further shown that Vision-Language Models (VLMs) may produce geo-economic biased responses even when interpreting correctly designed charts \cite{mahbub2025charts}. Therefore, when VLMs are susceptible to misleading visual designs, the consequences can be substantial, including poor financial judgments, regulatory errors, and the amplification of misinformation. 
Figure~\ref{fig:teaser_mitigation} illustrates a representative example in which a simple inverted-axis manipulation causes a state-of-the-art VLM to draw the opposite conclusion from the same underlying data.
As VLMs become increasingly integrated into data analysis workflows, ensuring their ability to interpret visualizations faithfully is essential and urgent, particularly because their outputs can influence real-world decisions \cite{fisher2025biased}.

VLMs have shown remarkable promise in tasks such as chart summarization \cite{kantharaj2022chart}, question answering \cite{hoque2022chart}, visualization generation \cite{mahbub2025charts}, and improving chart accessibility for individuals with visual impairments or low visual literacy \cite{choe2024enhancing, gorniak2024vizability}. Given their growing role in analysis and accessibility, it is essential that VLMs can accurately interpret data, even when visualizations are designed to deceive. Previous research has explored VLMs’ capacity to characterize misleading charts, but often under idealized conditions where prompts explicitly hint at the deception \cite{alexander2024can, lo2024good}. In everyday use, however, people usually engage with visualizations through neutral prompts, unaware of potential misrepresentation \cite{hoque2022chart}. 

A key challenge in studying chart deception in VLMs is evaluation design. Existing analyses often compare responses between faithful and misleading charts using aggregate accuracy or statistical shifts~\cite{mahbub2025charts}. However, such differences do not necessarily reflect deception-induced failures, since VLMs already exhibit baseline reasoning errors on standard chart understanding tasks~\cite{masry2025chartqapro}. Moreover, without controlling for underlying data and task characteristics, it is difficult to attribute response changes specifically to deceptive visual manipulations.  Distinguishing ordinary reasoning failures from errors specifically caused by deceptive visual designs is therefore essential for reliably measuring deception susceptibility.

Beyond evaluation, improving robustness to misleading visualizations presents an equally important challenge. Existing approaches often rely on explicitly prompting models to inspect charts for predefined deception types~\cite{das2025misvisfix, chen2025unmasking}, implicitly assuming that users already suspect deception. In practice, however, chart interpretation is typically cue-free. We therefore investigate whether structured intermediate grounding at inference time can improve robustness to misleading visualizations without requiring deception-aware prompts or model retraining. This motivates mitigation strategies that remain practical for real-world deployment.

In this work, we present a comprehensive study of how misleading visualizations influence VLM reasoning in cue-free settings. More specifically, we investigate three key research questions:\\
\noindent \textbf{RQ1:} To what extent do subtle misleading chart designs influence VLM responses, and how does susceptibility vary across different deception types and models sizes?\\ 
\noindent \textbf{RQ2:} How can deception-induced reasoning failures be distinguished from ordinary chart reasoning errors?\\
\noindent \textbf{RQ3:} Can inference-time mitigation approaches improve the robustness of these VLMs to misleading chart designs?

To answer these questions, our work makes three main contributions: \textbf{(1)} 
We introduce VisDeception, a benchmark of 1,600 paired faithful and misleading charts spanning eight categories of deceptive visualization tactics, together with a metric for measuring deception-induced reasoning failures while accounting for baseline chart reasoning errors. \textbf{(2)} 
We present an inference-time multi-agent grounding framework that improves robustness to misleading visualizations without requiring deception-aware prompts or model retraining. \textbf{(3)}~We conduct a large-scale evaluation of 10 state-of-the-art proprietary and open-source VLMs across more than 48,000 responses, showing that even advanced models remain highly vulnerable to deceptive visual manipulations, while our mitigation framework substantially improves robustness across several leading systems. To facilitate future research on trustworthy chart understanding, we publicly release our benchmark, prompts, and code at \href{https://github.com/vis-nlp/visDeception}{https://github.com/vis-nlp/visDeception}.

A preliminary short-paper version of this work appeared in IEEE VIS 2025 \cite{mahbub2025perils}. This article substantially extends the work in three ways. First, we expand the benchmark with additional deception categories, finer-grained subtypes, and broader coverage of misleading visualization tactics. Second, we introduce the Deception Score, a paired evaluation metric designed to isolate deception-induced reasoning failures from baseline chart reasoning errors. Third, we present an inference-time mitigation framework and demonstrate its effectiveness across a large suite of modern VLMs.

\section{Related Work}
\subsection{VLMs for Visualizations}
Recent advances in vision-language models (VLMs) have substantially improved machine understanding and generation of visualizations~\cite{islam2024large}. Prior work has introduced a wide range of visualization-centered tasks, including chart summarization and captioning~\cite{chart-to-text-acl, tang2023vistext}, chart question answering~\cite{masry-etal-2022-chartqa, open-CQA, lee2022pix2struct}, chart fact-checking~\cite{akhtar-etal-2023-reading, akhtar2023chartcheck}, multimodal data storytelling~\cite{shen2024datastory, islam-etal-2024-datanarrative, fu2025dataweaver}, and visualization generation from natural language~\cite{rahman-etal-2025-text2vis, shuai2025deepvis}. These developments increasingly position VLMs as tools for data analysis, communication, and decision support.

Existing VLMs for visualization tasks can broadly be grouped into general-purpose multimodal models and chart-specialized models. General-purpose proprietary models, such as GPT-4o~\cite{achiam2023gpt}, Claude~\cite{claude}, and Gemini~\cite{comanici2025gemini}, currently achieve state-of-the-art performance on many chart understanding benchmarks~\cite{wang2024charxiv}, often outperforming open-source alternatives such as LLaVA~\cite{li2024llava}, InternVL~\cite{internvl}, and Qwen~\cite{yang2025qwen3}. At the same time, the adaptability of open-source models has enabled the development of chart-specific systems, including UniChart~\cite{masry2023unichartuniversalvisionlanguagepretrained}, ChartInstruct~\cite{masry-etal-2024-chartinstruct}, ChartGemma~\cite{masry-etal-2025-chartgemma}, and BigChartR1~\cite{masry2025bigcharts}, which demonstrate strong performance on chart reasoning benchmarks~\cite{masry2022chartqa, akhtar2023chartcheck, kantharaj2022chart}.

Despite these advances, recent studies show that VLMs remain vulnerable to hallucinations, factual inconsistencies, reasoning failures, and socially biased interpretations when analyzing visualizations~\cite{islam2024large, huang2024pixels, mahbub2025charts}. However, prior work has not systematically examined how VLMs respond to misleading chart designs or whether such vulnerabilities can be mitigated at inference time, which we address in this work.

\subsection{Misleading Visual Designs}

The potential for charts to mislead has been recognized since the 1950s, with classic works like \emph{``How to Lie with Statistics''}~\cite{huff2023lie} illustrating how visual and statistical representations can distort information and contribute to misinformation. Researchers have examined this issue from multiple perspectives. Some studies have investigated deception strategies in common chart types, i.e., graphs, lines, and pie, 
through user studies assessing their impact on interpretation and decision-making~\cite{pandey2015deceptive, lauer2020deceptive}. Lan \textit{et al.} conducted five focus group studies to investigate the underlying reasons behind visualization design flaws \cite{lan2024came}. Ge \textit{et al.} proposed a definition of visualization literacy that includes the ability to recognize visual deception, arguing that this skill is a vital component of overall visual literacy \cite{ge2023calvi}.  Lisnic \textit{et al.} \cite{lisnic2023misleading, 10.1145/3613904.3642448} analyzed various tactics used to spread misleading narrative using deceptive visualization in social media. Some studies have focused on very specific aspects of visual deception tactics, such as axis truncation \cite{correll2020truncating, long2024cut}.  While the effects of misleading charts on human interpretation have been widely studied, their impact on VLMs remains largely unexplored.

\begin{figure*}[t!]
    \centering
    \includegraphics[width = \textwidth]{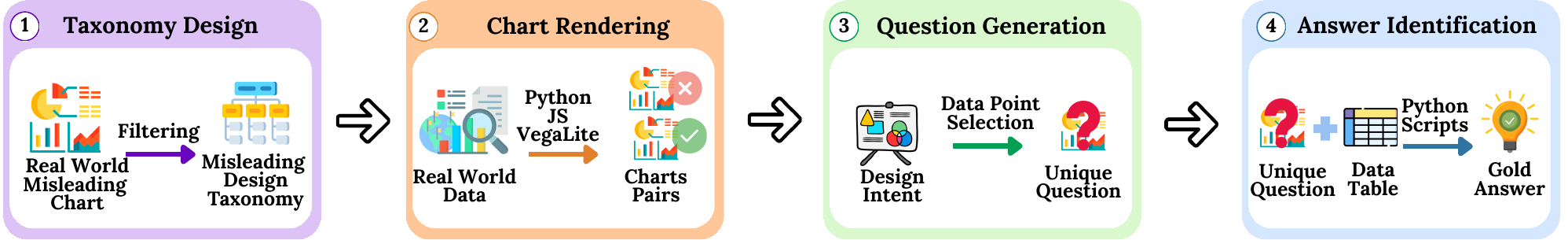} 
\caption{
Overview of our benchmark construction process: \textbf{(1)} We first design a taxonomy of common misleading chart designs looking into real world cases, \textbf{(2)} Using different systems, we render the charts where each sample contains an original and a misleading chart, \textbf{(3)} Taking into account the design intent behind the particular misleading chart, we generate question for each sample, \textbf{(4)} Using the corresponding data table of each sample, we identify the gold answer for the question.
}    
     \label{fig:methodology}
\end{figure*}

\subsection{VLMs and Misleading Visualizations}

Recent work has begun exploring whether VLMs can recognize misleading visualizations~\cite{lo2024good, alexander2024can, das2025misvisfix}. Lo \textit{et al.}~\cite{lo2024good} first studied misleading chart detection using real-world examples, while Alexander \textit{et al.}~\cite{alexander2024can} examined misleading visualizations in social media posts. However, these studies frame the task as \emph{deception detection}, where prompts or surrounding context often imply that the chart may be misleading. In contrast, real-world chart understanding is typically cue-free: users ask analytical questions without suspecting deception. Our work instead studies whether misleading designs silently alter VLM reasoning under such naturalistic settings.

More recently, research has explored mitigation strategies for misleading visualizations. Fan \textit{et al.}~\cite{fan2022annotating} proposed annotation-based interventions to help human readers better interpret deceptive line charts, but their goal was improving human visual literacy rather than improving VLM robustness. Misleading ChartQA~\cite{chen2025unmasking} introduced a benchmark for evaluating VLMs on misleading chart question answering; however, it lacks paired faithful counterparts for the same charts, making it difficult to isolate deception-induced errors from baseline reasoning failures. Its mitigation approach also relies on explicitly prompting predefined deception categories. MisVisFix~\cite{das2025misvisfix} presents an interactive system for detecting and correcting misleading visualizations using deception-aware prompting and predefined taxonomies. In contrast, our work studies how deceptive designs affect VLM reasoning itself and introduces an inference-time mitigation framework that improves robustness without requiring explicit deception-aware prompts or prior knowledge of specific deception types.

\section{The \benchmark Benchmark}

We introduce \benchmark, a controlled benchmark designed to systematically evaluate whether and how misleading chart designs affect the performance of VLMs. Our benchmark construction method consists of four stages (Figure~\ref{fig:methodology}). First, we develop a taxonomy of common deceptive visualization strategies grounded in prior work on misleading chart design. Second, we construct a paired chart corpus containing both faithful and misleading versions of each chart generated from the same underlying data. Third, for each deception type, we identify the intended perceptual distortion (e.g., exaggerating differences or reversing trends) and design targeted questions that directly probe susceptibility to that misleading visual effect. Finally, we compute gold answers programmatically from the underlying data tables to ensure consistency. Together, this pipeline enables controlled and fine-grained evaluation of deception-induced reasoning failures in VLMs.

\subsection{Taxonomy Design}

\begin{table}[t]
\centering
\renewcommand{\arraystretch}{1.4}
\caption{
Taxonomy of misleading chart designs included in VisDeception. For each category, we list the corresponding subcategories and chart types. Numbers in parentheses denote the number of paired chart instances per subcategory.
}
\label{tab:taxonomy_breakdown}
\resizebox{\columnwidth}{!}{%
\begin{tabular}{p{1.32cm}p{2.1cm}p{4.7cm}p{1.5cm}}
\toprule[1pt]
\textbf{Category} & \textbf{Major Category} & \textbf{Subcategories} & \textbf{Chart Type} \\ \hline

 & \cellcolor[HTML]{EFEFEF}\texttt{\textbf{Truncated Axis}}
 & \cellcolor[HTML]{EFEFEF}Middle Truncation (25), Vertical Truncation (50), Horizontal Truncation (25)
 & \cellcolor[HTML]{EFEFEF}Bar \\ \cline{2-4}

 & \texttt{\textbf{Aspect Ratio}}
 & Expansion (50), Compression (50)
 & Line \\ \cline{2-4}

 & \cellcolor[HTML]{EFEFEF}\texttt{\textbf{Dual Axis}}
 & \cellcolor[HTML]{EFEFEF}Dual-axis (100)
 & \cellcolor[HTML]{EFEFEF}Line \\ \cline{2-4}

\multirow{-4}{=}{Axis\\ Manipulation}
 & \texttt{\textbf{Inverted Axis}}
 & Inverted Axis (100)
 & Line \\ \hline

 & \cellcolor[HTML]{EFEFEF}\texttt{\textbf{Distorted Projection}}
 & \cellcolor[HTML]{EFEFEF}Pie Charts (75), Donut Charts (25)
 & \cellcolor[HTML]{EFEFEF}Pie, Donut \\ \cline{2-4}

 & \texttt{\textbf{Data-Visual Disproportion}}
 & Data-Visual Disproportion (100)
 & Bubble, Scatter \\ \cline{2-4}

 & \cellcolor[HTML]{EFEFEF}\texttt{\textbf{Inappropriate Encoding}}
 & \cellcolor[HTML]{EFEFEF}Inappropriate Categorical Encoding (50), Inappropriate Continuous Encoding (50)
 & \cellcolor[HTML]{EFEFEF}Line, Bar \\ \cline{2-4}

\multirow{-4}{=}{Misleading Encoding}
 & \texttt{\textbf{Inappropriate Color Coding}}
 & Close Colors (38), Rainbow Colors (25), Disordered Colors (37)
 & Choropleth Map \\

\toprule[1pt]
\end{tabular}%
}
\end{table}
\subsubsection{Benchmark Construction Principles}

A central challenge in constructing a benchmark for chart deception is isolating the effect of deceptive visual encodings from other factors that influence interpretation. 
Building on insights from our earlier study~\cite{mahbub2025perils} and an analysis of real-world misleading charts, we constructed \benchmark around four principles that emphasize controlled evaluation, causal attribution, and broad coverage of deceptive visualization practices.

\noindent\textbf{P1. Design-Based Deception Scope.}
Real-world misleading visualizations often combine visual manipulations with captions, annotations, framing, and narrative context~\cite{lo2022misinformed, lisnic2023misleading}. To enable controlled evaluation, we focus on \textit{design-based} deception, where the underlying data remain unchanged but visual encodings distort interpretation. Examples include axis reversals that invert perceived trends and scale manipulations that exaggerate small differences. These designs have been widely recognized as common and influential sources of misleading visual interpretation~\cite{lauer2020deceptive, pandey2015deceptive, szafir2018good}.

\noindent\textbf{P2. Consistent Question and Evaluation Design.} 
To enable systematic comparison across deception types, we restricted the benchmark to chart designs that are compatible with generalized survey-style analytical questions with programmatically verifiable gold answers, following the methodology of Lauer \textit{et al.}~\cite{lauer2020deceptive}. This design allows all chart pairs to be evaluated using consistent prompting and a common response format, enabling controlled measurement of deception-induced reasoning failures across models. For example, a comparative question about relative magnitudes in a truncated-axis chart can directly reveal whether the altered visual encoding changes model judgment.

\noindent\textbf{P3. Controlled Paired Comparison.}
A key design principle of our benchmark is controlled paired comparison. Following prior work on deceptive visualization evaluation~\cite{lauer2020deceptive}, each misleading chart is paired with a corresponding faithful version generated from the same underlying data. The misleading variant differs from the faithful chart by exactly one intentionally deceptive intervention. 
This construction controls for data content, chart semantics, visual complexity, and task difficulty, enabling changes in model behavior to be attributed directly to the deceptive manipulation itself. 

\noindent\textbf{P4. Broad Taxonomy Coverage.} 
Based on these principles, we organized the benchmark into eight major categories of deceptive visualization strategies, several of which are further divided into subcategories capturing different manifestations of the same high-level deception mechanism. Together, these categories span a broad spectrum of manipulations that alter the perceived magnitude, trend, relation, ordering, or salience of data without modifying the underlying values.

\begin{figure*}[t!]
  \centering
  \includegraphics[width=\linewidth, alt={Overview of the eight misleading chart designs studied.}]{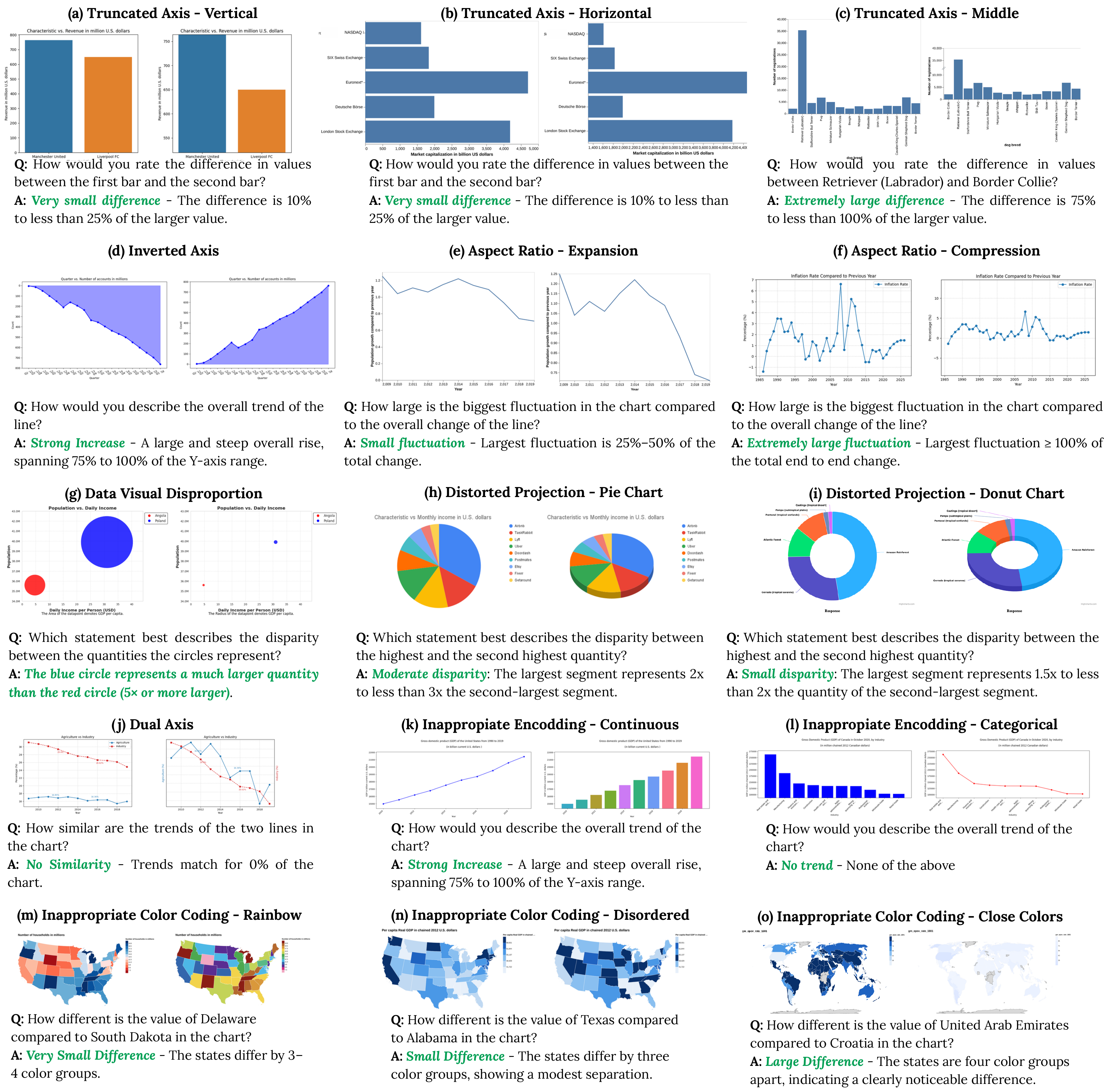}
  \caption{Overview of the eight misleading chart designs studied, along with the subtypes of each of the major types. For each category, the chart on the left is the original chart and the chart on the right is its misleading counterpart. The Misleading design type is given on top of each pair, with the sub-type after the hyphen if applicable. }
  \label{fig:teaser}
\end{figure*}

\subsubsection{Misleading Visualization Types}  
We now describe each misleading category included in the benchmark.

\noindent \textbf{Aspect Ratio Distortion.} Aspect ratio distortion arises when the height-to-width ratio of a chart
is adjusted to change the perceived steepness of trends. Although the underlying data values remain unchanged, modifying the chart geometry can make the same trend appear either more dramatic or more subtle, a perceptual effect discussed in prior work on deceptive visualization and aspect-ratio selection~\cite{lauer2020deceptive, heer2006multi, cleveland1993visualizing}. We consider two subcategories in this group: \textit{Expansion}, where the chart is stretched to amplify perceived differences, and \textit{Compression}, where the chart is compressed to visually flatten variation. Examples of these two subcategories are shown in Figure~\ref{fig:teaser}(e) and Figure~\ref{fig:teaser}(f), respectively.

\noindent \textbf{Data--Visual Disproportion.}
Visual magnitude can become misleading when it is not proportional to the underlying data magnitude. A common example is a bubble chart in which bubble size is scaled by \textit{radius} rather than \textit{area}, causing the visual encoding to disproportionately exaggerate perceived differences~\cite{pandey2015deceptive}. To capture varying levels of structural complexity, we include both \textit{two-point} examples, which involve simple pairwise comparisons (Figure~\ref{fig:teaser}(g)), and \textit{multi-point} examples, which contain several bubbles. In both settings, the misleading effect arises from a mismatch between the encoded visual size and the actual data values. 

\noindent \textbf{Distorted Projection.}
Perspective-based or 3D chart designs can distort perceived relative sizes by altering the visual prominence of chart elements. Prior works~\cite{szafir2018good,lauer2020deceptive} have shown that in charts such as pie and donut charts, 3D projection can make front-facing segments appear larger and back-facing segments appear smaller, even when the underlying proportions remain unchanged. To capture this form of visual distortion, we include two subcategories: \textit{Pie Charts}, where 3D perspective affects slice comparison in traditional pie charts, and \textit{Donut Charts}, where similar projection effects are applied to ring-shaped charts (Figure~\ref{fig:teaser}(h) and Figure~\ref{fig:teaser}(i)). 

\noindent \textbf{Inappropriate Encoding.}
We include cases where the chosen chart type or visual encoding is poorly matched to the underlying data type, following prior work that identifies such mismatches as a source of misleading visual interpretation \cite{lisnic2023misleading}. For example, using a line chart for categorical data may falsely imply continuity or a meaningful trend, whereas applying a bar-like encoding to continuous variation may obscure gradual changes. We consider two subcategories: \textit{Inappropriate Categorical Encoding}, where continuous visual forms are used for categorical data, and \textit{Inappropriate Continuous Encoding}, where categorical-style encodings are used for data that are more naturally interpreted as continuous (Figure \ref{fig:teaser}(k),(l)). 

\noindent \textbf{Dual Axis.}
The dual-axis deceptive category covers misleading uses of axes that distort the perceived relationship between variables as shown in the works of Lisnic et al.~\cite{lisnic2023misleading}. A representative example is the dual-axis chart shown in figure \ref{fig:teaser}(j), where different scales are placed on separate axes in a way that can create an artificial impression of alignment or correlation.

\noindent \textbf{Inappropriate Color Coding.}
Misleading color choices are especially consequential in choropleth maps and other value-encoded visualizations, where color directly shapes how viewers compare magnitudes across regions. Non-monotonic palettes, rainbow schemes~\cite{borland2007rainbow}, disordered scales, and overly similar colors can obscure ordered value relationships, introduce artificial visual contrasts, or unintentionally emphasize and suppress particular regions~\cite{harrower2003colorbrewer}. We therefore include both \textit{state-level} and \textit{country-level} map settings, together with three misleading color subcategories: \textit{Rainbow}, shown in Figure~\ref{fig:teaser}(m), where hue variation introduces artificial visual contrast; \textit{Disordered}, shown in Figure~\ref{fig:teaser}(n), where the color progression does not follow a meaningful perceptual order; and \textit{Close Colors}, shown in Figure~\ref{fig:teaser}(o), where overly similar shades make important differences difficult to distinguish.

\noindent \textbf{Inverted Axis.}
The inverted-axis category reverses the direction of an axis so that increasing values appear to decrease, or vice versa. Such inversion can fundamentally alter the perceived narrative of a chart by flipping trend direction or reversing the apparent ordering of values. As shown in figure \ref{fig:teaser}(d), the misleading effect comes from this \textit{axis reversal}~\cite{pandey2015deceptive}, which changes the visual interpretation of the chart while leaving the data values unchanged.

\noindent \textbf{Truncated Axis.}
Truncated axes use non-zero baselines or broken scales to visually amplify differences between values, particularly in bar charts, a practice that has been discussed in prior work on deceptive visualization design~\cite{pandey2015deceptive, lauer2020deceptive}. Even when the underlying numerical difference is relatively small, axis truncation can make one value appear disproportionately larger than another, thereby distorting comparative judgment. To capture these variations, we include several subcategories: \textit{Vertical Truncation} (Figure~\ref{fig:teaser}(a)), where the baseline is shifted upward from zero; \textit{Middle Truncation} (Figure~\ref{fig:teaser}(c)), where a portion of the axis is visually omitted; and \textit{Horizontal Truncation} (Figure~\ref{fig:teaser}(b)), where truncation is applied along the horizontal axis.

\subsection{Chart Corpus Creation}

In total, \benchmark contains \textbf{8 major deception categories}, which can be further divided into 15 sub-categories in total, each major deception categories with \textbf{100 paired instances}, resulting in \textbf{800 original charts}, \textbf{800 misleading charts}, and \textbf{1,600 chart images} overall. Each misleading chart is paired with a corresponding faithful counterpart generated from the same underlying data, enabling controlled evaluation of deception-induced reasoning failures while isolating the effect of the misleading visual manipulation itself. 

The underlying data tables are stored in CSV or JSON format and are derived from the ChartQA benchmark, which was collected from real-world sources~\cite{masry2022chartqa}.  Using real-world data while controlling chart generation enables realistic chart content while ensuring that deceptive visual interventions can be introduced systematically and reproducibly.

To ensure diversity in visual appearance and chart structure while preserving consistency in benchmark construction, we employed multiple rendering frameworks, including \textbf{Matplotlib}~\cite{Hunter:2007}, \textbf{Vega-Lite}~\cite{2017-vega-lite}, and \textbf{Highcharts.js}~\cite{highcharts2026github}. Matplotlib and Vega-Lite were used for most chart categories involving structured visual transformations such as aspect-ratio distortion, axis truncation, and encoding manipulations. Vega-Lite was additionally used for choropleth visualizations because of its stronger support for map-based color encodings.

A small number of deception types, such as middle-axis truncation (Figure~\ref{fig:teaser}(c)) and certain inverted-axis variants (Figure~\ref{fig:teaser}(d)), required manual post-processing because they were not directly supported by Matplotlib or Vega-Lite. In these cases, manual edits were restricted strictly to the targeted deceptive intervention in order to preserve controlled paired comparison.

\subsection{Question-Answer Pair Collection}
\begin{figure}[t]
    \centering
    \includegraphics[scale = 0.5]{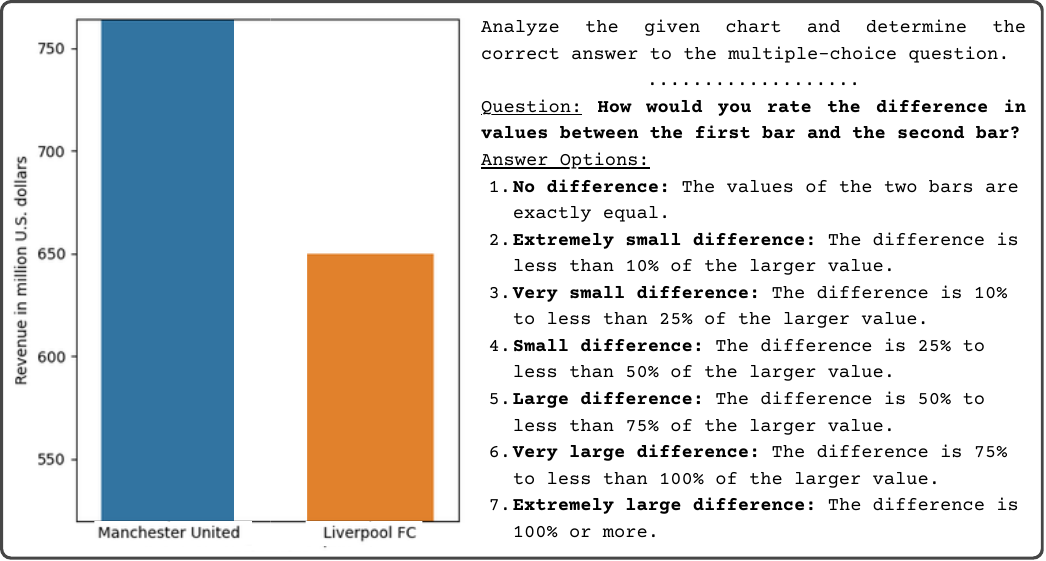} 
\caption{
Example question instance for the truncated-axis deception category. The truncated y-axis visually amplifies the difference between the bars, and the associated question tests whether the model is misled by this distortion.
 }
\label{fig:question}

\end{figure}

For each deception type, we designed targeted questions that directly probed the intended misleading visual effect. All questions used a standardized 7-point Likert-style response scale following prior work~\cite{mahbub2025perils}, enabling consistent comparison across chart types, deception categories, and VLMs. Each question was applied to both the faithful and misleading versions of the same chart. Since the underlying data remain unchanged across the pair, a robust VLM should ideally produce identical responses for both versions; any response shift can therefore be attributed to the misleading visual design.

 \noindent\textbf{Design Intent Identification.} 
Building on the taxonomy described above, the first step in question construction was identifying the primary perceptual distortion associated with each deception category. For example, Figure~\ref{fig:question} shows a truncated-axis chart comparing the revenues of `Manchester United' and `Liverpool FC'. Although the underlying values differ only moderately, the truncated axis visually exaggerates the gap between the two bars. The corresponding design intent is therefore to amplify the perceived difference between the values. More generally, the deception categories in our taxonomy are associated with distinct perceptual distortions, such as exaggerating magnitude differences, reversing perceived trends, obscuring value relationships, or creating a false impression of correlation.

\noindent\textbf{Question Design.}
Questions were tailored to the design intent of each deception category. For example, in truncated-axis charts, the goal was to exaggerate or suppress perceived differences between chart elements. Accordingly, we prompted the VLM with the question: ``\textit{How would you rate the difference in values between the first bar and the second bar?}'' This required the model to reason about the perceived visual difference between the two values. Following this principle, we developed one generalized question template for each deception type. All questions were presented in multiple-choice format with seven ordered response options. For charts containing two data points, the same generalized question was used across all instances. For charts with multiple data points, we identified the element most affected by the deceptive manipulation and selected an appropriate comparison target from the remaining chart elements.

\noindent\textbf{Gold answer identification:}
After generating each question, we computed the gold answer directly from the underlying data table rather than from the rendered visualization. This ensures that the reference answer reflects the true data values and is unaffected by the misleading visual design. In the example shown in Figure~\ref{fig:question}, we retrieved the values for `Manchester United' and `Liverpool FC' from the data table and used a Python script to compute the correct response for the corresponding Likert-scale question. The resulting difference is approximately 15\%, which maps to \textit{Option 3}. We repeated this procedure for all deception types. Because each faithful--misleading pair shares the same underlying data, both charts are assigned the same gold answer.

\section{Methodology}

This section describes the experimental setup and evaluation metrics used to quantify deception-induced reasoning failures in VLMs.

\subsection{Model Selection}

We selected VLMs spanning different architectures, parameter scales, and deployment settings in order to obtain a representative evaluation of state-of-the-art chart understanding systems. Our evaluation includes both proprietary and open-source models, allowing us to analyze whether robustness to misleading visualizations varies across model families, model sizes, and training paradigms. Among proprietary systems, we evaluated GPT-4o~\cite{achiam2023gpt}, Claude Haiku 4.5~\cite{claude}, Gemini 2.5 Flash~\cite{comanici2025gemini}, and Gemini 2.5 Pro~\cite{comanici2025gemini}. For open-source models, we selected multiple model families and parameter scales, including Gemma 3 12B IT~\cite{team2024gemma}, InternVL 3.5 2B, 4B, and 8B~\cite{internvl}, Pixtral 12B~\cite{agrawal2024pixtral}, and Qwen3-VL 2B, 4B, and 8B~\cite{yang2025qwen3}. Including multiple parameter scales within the same family enabled controlled analysis of whether susceptibility to misleading visualizations changes as model capacity increases.

\subsection{Quantifying Deception}
\label{sec:deception_metric}

A central challenge in evaluating misleading visualizations is distinguishing deception-induced reasoning failures from ordinary chart-understanding errors. Existing approaches often rely on aggregate accuracy or raw response differences between faithful and misleading charts~\cite{mahbub2025charts, lauer2020deceptive}. However, response changes alone do not necessarily indicate deception because VLMs may already produce incorrect answers on the original chart. A meaningful deception metric must therefore isolate the additional error introduced specifically by the misleading visual design.

Our paired benchmark enables such measurement. Each faithful–misleading chart pair shares the same underlying data, question, and gold answer, differing only in the deceptive visual intervention. We therefore define the \emph{Deception Score}, which measures the increase in ordinal prediction error induced by the misleading chart relative to its faithful counterpart. Formally, for a model evaluated on $N$ paired chart samples, we define:

\begin{equation}
\mathrm{DeceptionScore}
=
\frac{1}{N}
\sum_{i=1}^{N}
\left[
D(M_i, G_i) - D(O_i, G_i)
\right],
\label{eq:deception}
\end{equation}

\noindent where $G_i$ denotes the gold answer for chart pair $i$, $O_i$ denotes the model prediction for the faithful chart, $M_i$ denotes the prediction for the corresponding misleading chart, and $D(\cdot)$ represents the ordinal distance from the gold answer.

Because all questions use a seven-point ordered response scale (Figure~\ref{fig:question}), we compute ordinal distance as the absolute difference between option indices:

\begin{equation}
D(O_i, G_i) = |O_i - G_i|,
\qquad
D(M_i, G_i) = |M_i - G_i|.
\label{eq:ordinal_distance}
\end{equation}

\noindent For example, if $G_i=3$, $O_i=4$, and $M_i=6$, then the faithful-chart error is $1$, the misleading-chart error is $3$, and the resulting deception contribution is $3-1=2$.

The \emph{Deception Score} has a direct interpretation. Positive values indicate that the misleading chart increases model error relative to the faithful chart, zero indicates no additional deception-induced error, and negative values indicate that the model moves closer to the gold answer under the misleading visualization. Because responses lie on a seven-point scale, each instance-level contribution ranges from $-6$ to $+6$.

Importantly, the \emph{Deception Score} differs from simple response-shift measures such as $D(M_i,O_i)$. By anchoring both predictions to the same gold answer, the proposed metric quantifies increased task error rather than answer instability alone. We therefore report the \emph{Deception Score} together with faithful-chart and misleading-chart distances in order to distinguish baseline chart-understanding limitations from deception-induced reasoning failures.

\begin{figure*}[t]
    \centering
    \includegraphics[width=\linewidth]{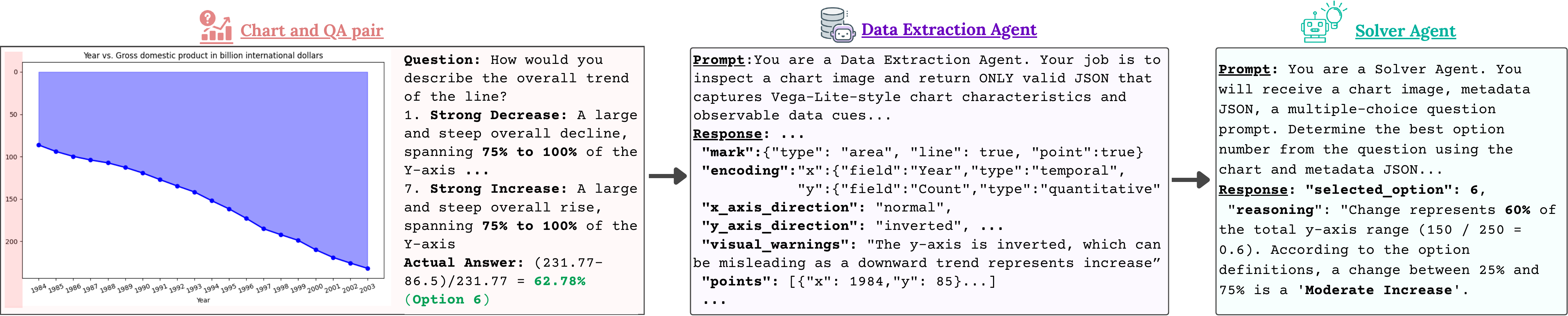} 
\caption{Inference time multi-agent mitigation framework. A Data Extraction Agent first converts the chart into a structured representation capturing chart semantics, data values, and key chart properties. A Solver Agent then reasons over both the chart and the extracted representation, grounding its answer in chart structure rather than visual appearance alone.}
     \label{fig:mitigatingDeception}
\end{figure*}
\subsection{Mitigation Approach}

To improve VLM robustness against misleading visualizations, we develop a multi-agent inference-time framework as a strong baseline solution that encourages models to reason from structured chart semantics rather than relying primarily on potentially deceptive visual cues. Prior work has shown that multi-agent systems can improve reasoning quality and reduce undesirable model behaviors across a range of tasks, including visualization generation~\cite{islam-etal-2024-datanarrative, mahbub2026datareel} and bias mitigation in conversational settings~\cite{xu2025mitigating}. Motivated by these findings, we investigate whether a structured multi-agent pipeline can similarly reduce deception-induced reasoning failures in chart understanding.

Our approach is illustrated in Figure~\ref{fig:mitigatingDeception}. In the standard single-VLM setting, models often rely directly on visual appearance and therefore become susceptible to deceptive design choices such as inverted axes or truncated baselines. To mitigate this issue, we introduce a multi-agent framework consisting of a \textit{Data Extraction Agent} and a \textit{Solver Agent}.

\noindent\textbf{Stage 1: Structured Chart Extraction.}
The \textit{Data Extraction Agent} analyzes the chart image and generates a structured Vega-Lite-style JSON representation describing both the chart structure and the underlying data. To support comprehensive chart interpretation, the extraction prompt provides an example schema that combines standard Vega-Lite components (e.g., marks, data values, visual encoding specifications) with additional chart-specific metadata, including axis properties, scale ranges, and chart dimensionality (e.g., 2D vs. 3D). The prompt does not explicitly instruct the model to detect deception or identify specific misleading designs. Instead, it directs the model to deconstruct the chart and produce a detailed structural representation. The schema also includes an optional \texttt{visual\_warnings} field that allows the model to record noteworthy chart characteristics when appropriate. As illustrated in Figure~\ref{fig:mitigatingDeception} (middle), this field can capture open-ended observations about chart properties, including potentially misleading or perceptually ineffective design choices (e.g., inverted axes, truncated baselines, or unusual scaling behaviors). Requiring explicit reconstruction of chart structure encourages the model to reason about chart semantics and visual encodings before attempting downstream question answering.

\noindent\textbf{Stage 2: Grounded Reasoning.}
The \textit{Solver Agent} answers the chart question using both the original chart image and the structured metadata produced by the extraction stage. 
We retain the original image to avoid information loss arising from imperfect extraction, while the structured representation provides an additional grounding signal that exposes chart semantics and numerical relationships more explicitly. Providing this structured context encourages the model to reason from reconstructed chart semantics rather than relying primarily on potentially misleading visual appearance. 
As illustrated in Figure~\ref{fig:mitigatingDeception}, this additional grounding often corrects failures observed in the single-VLM setting, particularly for deception types involving axis manipulation and misleading visual scaling.

Overall, our hypothesis is that explicitly separating chart interpretation into structured extraction and downstream reasoning stages encourages more careful inspection of chart semantics and reduces susceptibility to deceptive visual patterns. Unlike prior mitigation approaches that require explicit deception-aware prompts or predefined deception taxonomies, our framework operates without directly informing the model that the visualization may be misleading. This design better reflects realistic deployment settings, where users typically query charts without suspecting deceptive visual manipulation.

 \begin{table*}[t]
\centering

\resizebox{\textwidth}{!}
{
\begin{tabular}{l|cc|cc|cc|cc|cc|cc|cc|cc}
\toprule[1pt]
\multicolumn{1}{l|}{} & \multicolumn{2}{c|}{\textbf{\begin{tabular}[c]{@{}c@{}}Aspect\\ Ratio\end{tabular}}} & \multicolumn{2}{c|}{\textbf{\begin{tabular}[c]{@{}c@{}}Distorted\\ Projection\end{tabular}}} & \multicolumn{2}{c|}{\textbf{\begin{tabular}[c]{@{}c@{}}Dual\\ Axis\end{tabular}}} & \multicolumn{2}{c|}{\textbf{\begin{tabular}[c]{@{}c@{}}Inappropriate \\ Encoding\end{tabular}}} & \multicolumn{2}{c|}{\textbf{\begin{tabular}[c]{@{}c@{}}Inappropriate\\ Color Coding\end{tabular}}} & \multicolumn{2}{c|}{\textbf{\begin{tabular}[c]{@{}c@{}}Inverted\\ Axis\end{tabular}}} & \multicolumn{2}{c|}{\textbf{\begin{tabular}[c]{@{}c@{}}Data Visual\\ Disproportion\end{tabular}}} & \multicolumn{2}{c}{\textbf{\begin{tabular}[c]{@{}c@{}}Truncated\\ Axis\end{tabular}}} \\ \cline{2-17}
\multicolumn{1}{l|}{\multirow{-2}{*}{\textbf{Models}}} & $z$, $p$ & $DS$ & $z$, $p$ & $DS$ & $z$, $p$ & $DS$ & $z$, $p$ & $DS$ & $z$, $p$ & $DS$ & $z$, $p$ & $DS$ & $z$, $p$ & $DS$ & $z$, $p$ & $DS$ \\ \toprule[1pt]
\multicolumn{17}{l}{\textit{\faLock \; Closed-Source Models}} \\
\rowcolor[HTML]{EFEFEF}
\multicolumn{1}{l|}{\cellcolor[HTML]{EFEFEF}GPT-4o} & -2.04, $\bm{0.03}$ & 0.02 & -0.57, 0.56 & 0.03 & -4.21, $\bm{2.3e^{-5}}$ & \textbf{0.25} & -0.69, 0.47 & 0.50 & -3.17, $\bm{1.4e^{-3}}$ & \textbf{0.41} & -3.12, $\bm{1.6e^{-3}}$ & 1.97 & -1.64, 0.10 & 0.46 & -3.87, $\bm{7.8e^{-5}}$ & 0.38 \\
\rowcolor[HTML]{EFEFEF}
\multicolumn{1}{l|}{\cellcolor[HTML]{EFEFEF}Claude-4.5-Haiku} & -0.31, 0.75 & -0.06 & -3.08, $\bm{1.1e^{-3}}$ & 0 & -3.83, $\bm{1.1e^{-4}}$ & 0.22 & -5.04, $\bm{4.2e^{-8}}$ & 0.32 & -2.71, $\bm{5.9e^{-3}}$ & 0.06 & -2.52, $\bm{0.01}$ & 2.19 & -2.46, $\bm{0.01}$ & 0.30 & -2.33, $\bm{0.02}$ & 0.34 \\
\rowcolor[HTML]{EFEFEF}
\multicolumn{1}{l|}{\cellcolor[HTML]{EFEFEF}Gemini-2.5-Pro} & -1.18, 0.23 & -0.07 & -0.31, 0.73 & -0.05 & -1.18, 0.21 & -0.05 & -2.11, $\bm{0.03}$ & 0.23 & -1.38, 0.16 & 0.37 & -0.20, 0.84 & 0.18 & -1.10, 0.26 & 0.19 & -0.26, 0.79 & -0.01 \\
\rowcolor[HTML]{EFEFEF}
\multicolumn{1}{l|}{\cellcolor[HTML]{EFEFEF}Gemini-2.5-Flash} & -1.22, 0.20 & 0.02 & -0.89, 0.34 & -0.06 & -0.71, 0.46 & 0.07 & -0.56, 0.53 & 0.02 & -2.24, $\bm{0.02}$ & 0.26 & -0.47, 0.63 & 0.39 & -0.91, 0.35 & 0.16 & -0.32, 0.74 & 0.01 \\
\midrule
\multicolumn{17}{l}{\textit{\faLockOpen \; Open-Source Models}} \\
\rowcolor[HTML]{DBF7FF}
\multicolumn{1}{l|}{\cellcolor[HTML]{DBF7FF}Gemma3-12B-IT} & -0.67, 0.49 & 0 & -2.06, $\bm{0.03}$ & \textbf{0.09} & -2.03, $\bm{0.03}$ & 0.01 & -4.56, $\bm{1.9e^{-6}}$ & 0.68 & -1.43, 0.15 & -0.15 & -3.12, $\bm{1.7e^{-3}}$ & 1.99 & -2.72, $\bm{5.1e^{-3}}$ & 0.05 & -2.14, $\bm{0.03}$ & -0.09 \\
\rowcolor[HTML]{DBF7FF}
\multicolumn{1}{l|}{\cellcolor[HTML]{DBF7FF}Pixtral-12B} & -1.58, 0.11 & \textbf{0.30} & -0.44, 0.64 & 0.06 & -0.25, 0.79 & -0.17 & -1.84, 0.06 & 0.35 & -0.53, 0.59 & 0.12 & -2.23, $\bm{0.02}$ & 2.43 & -2.40, $\bm{0.01}$ & \textbf{0.62} & -3.06, $\bm{1.9e^{-3}}$ & 0.53 \\
\rowcolor[HTML]{DBF7FF}
\multicolumn{1}{l|}{\cellcolor[HTML]{DBF7FF}Qwen3-VL-2B} & -0.42, 0.67 & -0.05 & -0.37, 0.71 & 0.07 & -- & 0 & -1.02, 0.30 & -0.17 & -0.53, 0.56 & -0.15 & -3.23, $\bm{8.4e^{-4}}$ & 2.57 & -4.17, $\bm{3.0e^{-6}}$ & -0.26 & -- & 0 \\
\rowcolor[HTML]{DBF7FF}
\multicolumn{1}{l|}{\cellcolor[HTML]{DBF7FF}Qwen3-VL-4B} & -0.70, 0.47 & 0.29 & -2.04, $\bm{0.04}$ & 0.08 & -0.71, 0.42 & -0.04 & -3.15, $\bm{7.6e^{-4}}$ & 0.34 & -3.58, $\bm{3.2e^{-4}}$ & 0.22 & -2.67, $\bm{5.6e^{-3}}$ & \textbf{2.61} & -2.22, $\bm{0.02}$ & 0.36 & -3.98, $\bm{6.0e^{-5}}$ & 0.63 \\
\rowcolor[HTML]{DBF7FF}
\multicolumn{1}{l|}{\cellcolor[HTML]{DBF7FF}Qwen3-VL-8B} & -1.54, 0.11 & 0.07 & -2.93, $\bm{2.9e^{-3}}$ & \textbf{0.09} & -4.02, $\bm{4.7e^{-5}}$ & 0.15 & -4.46, $\bm{4.8e^{-6}}$ & 1.58 & -2.71, $\bm{5.9e^{-3}}$ & 0.17 & -3.95, $\bm{6.7e^{-5}}$ & 2.09 & -0.69, 0.48 & 0.29 & -4.94, $\bm{5.4e^{-7}}$ & \textbf{0.66} \\
\rowcolor[HTML]{DBF7FF}
\multicolumn{1}{l|}{\cellcolor[HTML]{DBF7FF}Qwen3-VL-32B} & -0.87, 0.37 & 0.17 & -0.92, 0.35 & -0.01 & -5.91, $\bm{2.2e^{-9}}$ & 0.08 & -5.85, $\bm{1.1e^{-9}}$ & \textbf{1.80} & -3.22, $\bm{1.1e^{-3}}$ & 0.16 & -2.99, $\bm{2.7e^{-3}}$ & 2.11 & -0.87, 0.37 & 0.51 & -2.17, $\bm{0.03}$ & 0.15 \\
\toprule[1pt]
\end{tabular}
}

\caption{Results of the Wilcoxon signed-rank test and deception scores ($DS$) for each misleading chart category. Significant $p$-values ($p < 0.05$) are shown in \textbf{bold}. $DS$ (Deception Score) measures the average increase in error when viewing misleading charts compared to original charts. The highest $DS$ in each category is shown in \textbf{bold}.}
\label{tab:result_wilcoxon}
\end{table*}

 \section{Result Analysis}
We evaluate 10 VLMs on the 1,600 chart pairs in VisDeception, resulting in 16,000 responses under the single-agent setting. We then evaluate the same charts using the proposed multi-agent framework, producing an additional 16,000 responses (32,000 total). The following sections analyze statistical susceptibility to deception (RQ1), quantify deception-induced error using the proposed Deception Score (RQ2), and evaluate the effectiveness of the mitigation framework (RQ3).
\subsection{Statistical Analysis of Deception}
To assess whether misleading visualizations produce significant changes in model responses, we use paired Wilcoxon signed-rank tests, which are appropriate for matched chart pairs and do not assume normally distributed response differences. Table~\ref{tab:result_wilcoxon} summarizes the results across all models and deception categories. Overall, misleading chart designs significantly influence both closed-source and open-source VLMs, although the degree of susceptibility varies across models and deception types. Among closed-source models, Claude-3.5-Haiku is the most consistently deceived, showing significant response shifts for seven out of eight deception types, followed by GPT-4o, which is affected by five types. In contrast, Gemini models are more robust, with statistically significant effects limited to only a few categories. Open-source models generally exhibit broader and less stable vulnerabilities.  Both Qwen3-VL 4B and 8B are affected by six deception categories, while Gemma3-12B-IT also shows significant susceptibility across multiple settings. Importantly, robustness does not consistently improve with model scale. Larger models within the same family (e.g., Qwen3-VL-32B) often remain vulnerable to several deception types, suggesting that scaling alone is insufficient to mitigate misleading visual effects.

Across deception categories, \textit{Inverted Axis} emerges as the most consistently deceptive manipulation, producing statistically significant response shifts for nearly all evaluated models. This finding suggests that VLMs rely heavily on visually perceived trend direction and remain highly sensitive to axis orientation. \textit{Inappropriate Encoding}, \textit{Truncated Axis}, and \textit{Inappropriate Color Coding} also affect a broad range of models, indicating that VLMs remain vulnerable when visual encodings violate established visualization conventions. In contrast, \textit{Aspect Ratio Distortion} produces comparatively limited effectss, with significant effects appearing only for a single model.

Overall, these findings answer \textbf{RQ1} by demonstrating that subtle misleading visual designs can substantially alter VLM reasoning, but the magnitude and consistency of these effects depend strongly on both the underlying model architecture and the specific deception mechanism.

\subsection{Evaluation with Deception Score}

While the Wilcoxon analysis identifies statistically significant response shifts, it does not quantify how much those shifts increase task error.  We therefore evaluate all models using the proposed \emph{Deception Score}, which measures deception-induced error relative to performance on the corresponding faithful chart. 
Table~\ref{tab:single_multi_agent_deception} reports overall model-level scores, while Table~\ref{tab:result_wilcoxon} provides category-wise breakdowns.

The overall results reveal substantial differences in deception severity across architectures (Table~\ref{tab:single_multi_agent_deception}). Among the evaluated systems, Qwen3-VL-8B exhibits the highest overall vulnerability, while Gemini-2.5-Flash and Gemini-2.5-Pro achieve the lowest aggregate Deception Scores. More broadly, proprietary models tend to exhibit lower overall deception scores than most open-source systems, although no model is consistently robust across all deception categories.

A key observation is that the deception categories producing the largest statistical effects are also those producing the largest increases in task error (Table~\ref{tab:result_wilcoxon}). \textit{Inverted Axis} consistently yields the highest Deception Scores across models, indicating that reversing axis orientation fundamentally disrupts VLM interpretation of visual trends. \textit{Inappropriate Encoding} similarly produces substantial degradation, suggesting that VLMs strongly depend on learned associations between chart types and semantic meaning. Elevated scores are also observed for \textit{Truncated Axis} and \textit{Data--Visual Disproportion}, particularly among open-source models. In contrast, \textit{Aspect Ratio Distortion} and \textit{Distorted Projection} generally produce relatively small increases in task error, reinforcing the findings from the statistical analysis.

Importantly, the Deception Score reveals patterns that are not captured by statistical significance testing alone. Some categories produce statistically significant response shifts with only small increases in task error (e.g., GPT-4o on \textit{Aspect Ratio}: $p=0.03$, DS$=0.02$), while others substantially degrade reasoning quality (e.g., Qwen3-VL-32B on \textit{Inappropriate Encoding}: DS$=1.80$). Conversely, other categories substantially increase error despite failing to reach statistical significance (e.g., GPT-4o on \textit{Inappropriate Encoding}: $p=0.47$, DS$=0.50$), suggesting that significance testing alone may overlook meaningful degradation. 

Overall, these findings answer \textbf{RQ2} by showing that the proposed \emph{Deception Score} effectively separates deception-induced error from baseline chart-understanding error, providing a more informative measure of model vulnerability to misleading visualizations than statistical significance testing alone.

\subsection{Evaluation of Mitigation Approach}

We evaluate whether the proposed multi-agent framework improves robustness to misleading chart designs (\textbf{RQ3}) by introducing a structured chart extraction stage prior to question answering. Table~\ref{tab:single_multi_agent_deception} compares Deception Scores under direct VLM prompting and the proposed multi-agent setting.

Overall, the multi-agent framework reduces deception-induced error for most models, suggesting that grounding reasoning with structured chart metadata improves robustness to misleading visualizations. The gains are particularly pronounced for stronger models. Among proprietary systems, GPT-4o, Claude Haiku, and Gemini 2.5 Pro all show reductions in Deception Score, with Gemini 2.5 Pro improving from 0.099 to 0.024. Similar trends are observed for most open-source models. In particular, Qwen3-VL-32B decreases from 0.621 to 0.365, indicating that structured extraction is most effective when the underlying model can reliably recover and utilize chart semantics during downstream reasoning.

However, the benefits are not universal. Gemini 2.5 Flash and Qwen3-VL-2B exhibit increased Deception Scores under the multi-agent framework, suggesting that errors introduced during the extraction stage can propagate to downstream reasoning when the intermediate representation is incomplete or unreliable. These findings indicate that the effectiveness of the mitigation strategy depends not only on the reasoning capabilities of the solver agent but also on the quality of the extracted chart representation.

One important observation is that the Data Extraction Agent extraction prompt does not include any explicit hints about potential deceptive designs, unlike recent work~\cite{das2025misvisfix}. As a result, our prompt does not need to be manually updated when new deception types are introduced. Nevertheless, we observe that models often identify misleading visual elements, such as truncated axes, inverted scales, and exaggerated aspect ratios without being explicitly instructed to do so. The \textit{Solver Agent} then receives both the chart image and the extracted metadata as grounded context. This anchors its reasoning in explicit structural information about the chart, rather than relying solely on potentially misleading visual impressions. Figure~\ref{fig:mitigatingDeception} illustrates this process. In this example, Gemini 2.5 Pro incorrectly interprets an inverted-axis chart as showing a decreasing trend when answering the question directly from the image. However, the multi-agent approach resolves and it is able to reach the correct answer after using the extracted data.

\subsection{Evaluation of Mitigation on Major Categories}
Table~\ref{tab:gpt4o_qwen32b_single_multi_agent} reports the deception scores across major deception categories for GPT-4o and Qwen3-VL-32B, which are our best performing models among the closed and open source in terms of the mitigation approach. Among the evaluated models, GPT-4o shows the largest overall reduction in deception score under the multi-agent setup. We first observe that the multi-agent approach lowers the deception score across all major categories except Aspect Ratio, where there is a very slight increase. The highest deception score appears for Inverted Axis, with a single-agent score of 1.97. Under the multi-agent approach, this score decreases to 1.35, indicating a noticeable improvement. Inappropriate Color Coding also shows a substantial reduction of 0.48. Inappropriate Encoding, Data Visual Disproportion, Inappropriate Encoding, and Dual Axis show moderate decreases, while Distorted Projection exhibits only a slight decrease. Overall, these results suggest that the mitigation approach improves GPT-4o’s robustness across most deception categories, although the magnitude of improvement varies by deception type.

The category-level results also reveal important limitations. For Qwen3-VL-32B, the multi-agent setup increases Deception Scores for \textit{Truncated Axis} and \textit{Data--Visual Disproportion}, indicating that structured metadata extraction can occasionally introduce or amplify errors. These cases highlight that mitigation performance remains dependent on the fidelity of the extracted representation and is not uniformly beneficial across all deception mechanisms.

\begin{table}[t]
\centering
\small
\setlength{\tabcolsep}{5pt}
\renewcommand{\arraystretch}{1.1}

\resizebox{.49\textwidth}{!}{
\begin{tabular}{lccc}
\toprule[1pt]
\textbf{Model} & \textbf{Single Agent} & \textbf{Multi-Agent} & \textbf{Change} \\
\toprule[1pt]

\multicolumn{4}{l}{\textit{\faLock \; Closed-Source Models}} \\
\rowcolor[HTML]{EFEFEF}
GPT-4o & 0.502 & 0.255 & \textcolor{green!40!black}{0.247 $\downarrow$} \\
\rowcolor[HTML]{EFEFEF}
Claude 4.5 Haiku & 0.422 & 0.228 & \textcolor{green!40!black}{0.194 $\downarrow$} \\
\rowcolor[HTML]{EFEFEF}
Gemini 2.5 Pro & 0.099 & 0.024 & \textcolor{green!40!black}{0.075 $\downarrow$} \\
\rowcolor[HTML]{EFEFEF}
Gemini 2.5 Flash & 0.109 & 0.135 & \textcolor{red!80!black}{0.026 $\uparrow$} \\

\midrule
\multicolumn{4}{l}{\textit{\faLockOpen \; Open-Source Models}} \\
\rowcolor[HTML]{DBF7FF}
Gemma3 12B IT & 0.323 & 0.211 & \textcolor{green!40!black}{0.112 $\downarrow$} \\
\rowcolor[HTML]{DBF7FF}
Pixtral 12B & 0.530 & 0.395 & \textcolor{green!40!black}{0.135 $\downarrow$} \\
\rowcolor[HTML]{DBF7FF}
Qwen3-VL 2B & 0.251 & 0.275 & \textcolor{red!80!black}{0.024 $\uparrow$} \\
\rowcolor[HTML]{DBF7FF}
Qwen3-VL 4B & 0.561 & 0.339 & \textcolor{green!40!black}{0.222 $\downarrow$} \\
\rowcolor[HTML]{DBF7FF}
Qwen3-VL 8B & 0.638 & 0.513 & \textcolor{green!40!black}{0.125 $\downarrow$} \\
\rowcolor[HTML]{DBF7FF}
Qwen3-VL 32B & 0.621 & 0.365 & \textcolor{green!40!black}{0.256 $\downarrow$} \\
\bottomrule
\end{tabular}
}

\caption{Deception Scores: Single-Agent vs Multi-Agent Setup. The Change column indicates whether the deception score increases or decreases under the multi-agent setup.}
\label{tab:single_multi_agent_deception}
\end{table}

In relation to \textbf{RQ3:}, these findings suggest that inference-time mitigation approaches can meaningfully reduce the impact of misleading visualizations for certain models and categories, but they do not universally improve robustness in all models.

\subsection{Qualitative Analysis}
\begin{table}[t!]
\centering
\small
\setlength{\tabcolsep}{4pt}
\renewcommand{\arraystretch}{1.1}

\resizebox{.49\textwidth}{!}{
\begin{tabular}{l|ccc|ccc}
\toprule[1pt]
 & \multicolumn{3}{c|}{\textbf{GPT-4o}} & \multicolumn{3}{c}{\textbf{Qwen3-VL-32B}} \\
\textbf{Misleading Design} & \textbf{Single} & \textbf{Multi} & \textbf{Change} & \textbf{Single} & \textbf{Multi} & \textbf{Change} \\
\toprule[1pt]
\rowcolor[HTML]{EFEFEF}
Aspect Ratio & 0.02 & 0.05 & \textcolor{red!80!black}{0.03 $\uparrow$} & 0.17 & 0.11 & \textcolor{green!40!black}{0.06 $\downarrow$} \\
Distorted Projection & 0.03 & 0.02 & \textcolor{green!40!black}{0.01 $\downarrow$} & -0.01 & 0.06 & \textcolor{red!80!black}{0.07 $\uparrow$} \\
\rowcolor[HTML]{EFEFEF}
Dual Axis & 0.25 & 0.00 & \textcolor{green!40!black}{0.25 $\downarrow$} & 0.08 & -0.14 & \textcolor{green!40!black}{0.22 $\downarrow$} \\
Inappropriate Encoding & 0.50 & 0.24 & \textcolor{green!40!black}{0.26 $\downarrow$} & 1.80 & 1.01 & \textcolor{green!40!black}{0.79 $\downarrow$} \\
\rowcolor[HTML]{EFEFEF}
Inappropriate Color Coding & 0.41 & -0.07 & \textcolor{green!40!black}{0.48 $\downarrow$} & 0.16 & -0.02 & \textcolor{green!40!black}{0.18 $\downarrow$} \\
Inverted Axis & 1.97 & 1.35 & \textcolor{green!40!black}{0.62 $\downarrow$} & 2.11 & 0.95 & \textcolor{green!40!black}{1.16 $\downarrow$} \\
\rowcolor[HTML]{EFEFEF}
Data Visual Disproportion & 0.46 & 0.23 & \textcolor{green!40!black}{0.23 $\downarrow$} & 0.51 & 0.65 & \textcolor{red!80!black}{0.14 $\uparrow$} \\
Truncated Axis & 0.38 & 0.22 & \textcolor{green!40!black}{0.16 $\downarrow$} & 0.15 & 0.30 & \textcolor{red!80!black}{0.15 $\uparrow$} \\
\bottomrule
\end{tabular}
}

\caption{Deception scores for GPT-4o and Qwen3-VL-32B under single-agent and multi-agent setups. Change shows the direction and magnitude of shift from single to multi-agent. Here, less is better. 
}
\label{tab:gpt4o_qwen32b_single_multi_agent}
\end{table}

\begin{figure*}[t]
    \centering
    \includegraphics[scale = 0.272]{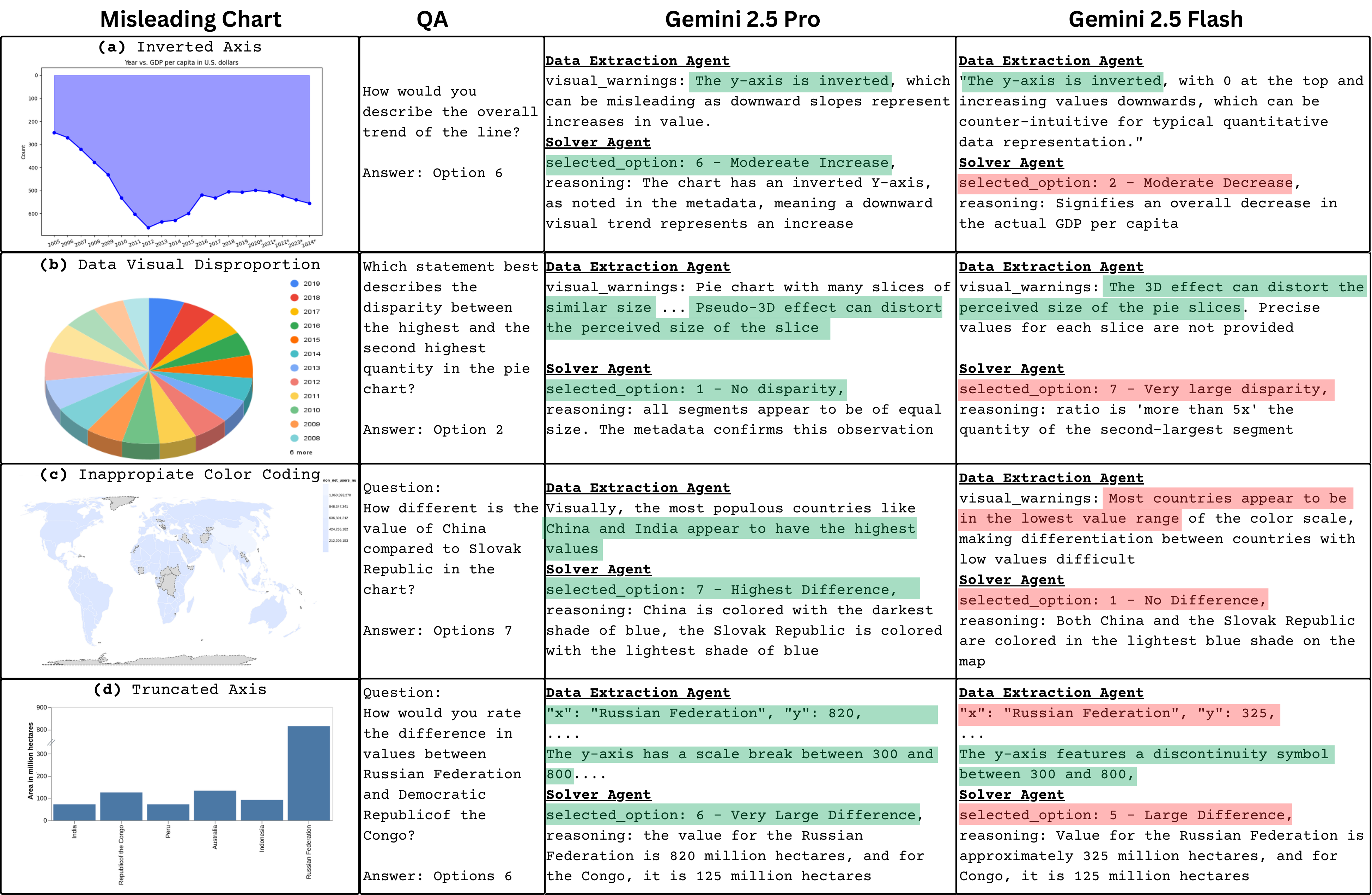} 
\caption{A qualitative analysis showing cases where mitigation approach has been success for the best model, Gemini 2.5 pro, but failed in case of its light-weight counterpart, Gemini 2.5 flash.
}
     \label{fig:qualitative}

\end{figure*}

To better understand why the mitigation framework succeeds for some models but fails for others, we qualitatively compared Gemini 2.5 Pro (the strongest mitigation performer) and Gemini 2.5 Flash across 50 misleading chart examples spanning different deception categories. Representative examples are shown in Figure~\ref{fig:qualitative}.

Our analysis reveals two recurring failure modes. The first arises when the \textit{Data Extraction Agent} correctly identifies a misleading chart property, but the \textit{Solver Agent} fails to incorporate this information during reasoning. Examples include inverted-axis and distorted-projection charts (Figure~\ref{fig:qualitative}a–b), where both models detect the misleading design, yet Gemini 2.5 Flash continues to rely primarily on visual appearance when generating its answer. For example, in Figure~\ref{fig:qualitative}a, both models identify the inverted axis, but only Gemini 2.5 Pro correctly utilizes its interpretation accordingly and recognizes the underlying increasing trend.

The second failure mode stems from errors during chart extraction itself (Figure~\ref{fig:qualitative}c–d). In these cases, the \textit{Data Extraction Agent} either misinterprets visual encodings or reconstructs incorrect chart values, leading the \textit{Solver Agent} to reason from faulty information. For example, in Figure~\ref{fig:qualitative}d, both models recognize the truncated axis, but Gemini 2.5 Flash incorrectly reconstructs the value of the Russian Federation, causing the downstream reasoning process to fail. Gemini 2.5 Pro, by contrast, accurately reconstructs the chart values and reaches the correct answer.

Overall, the qualitative analysis suggests that successful mitigation requires both accurate chart extraction and effective use of the extracted representation during downstream reasoning. These observations help explain why stronger models benefit more from the multi-agent framework: they are not only better at recovering chart semantics, but also more effective at leveraging structured representations when answering questions.

 \section{Discussion }

Our findings have implications for the evaluation, deployment, and future development of VLMs for chart understanding. We discuss the key lessons learned from this study, followed by limitations and directions for future work.

\textbf{Misleading visual designs remain a fundamental challenge for VLMs.} Across all evaluated models, misleading chart designs consistently altered model behavior, often increasing reasoning error even when the underlying data remained unchanged. While stronger models generally exhibited lower susceptibility, no model was uniformly robust across all deception categories. These findings suggest that current progress in chart understanding does not necessarily translate into robustness against misleading visual encodings. As VLMs become increasingly integrated into data analysis tools, educational platforms, and decision-support systems, robustness to misleading visualizations should be treated as a core evaluation dimension alongside answer accuracy.

\textbf{Studying visual deception in isolation remains important.} Real-world misinformation often combines deceptive visual designs with captions, annotations, and narrative framing. While such multimodal forms of deception are important, isolating visual design effects is a necessary first step toward understanding how VLMs interpret charts. By controlling the underlying data, questions, and answers, our benchmark allows changes in model behavior to be attributed directly to deceptive visual encodings. We view this level of experimental control as complementary to future studies of richer multimodal misinformation settings.

\textbf{Structured reasoning offers a promising direction for improving robustness.}
The success of the proposed multi-agent framework suggests that part of the vulnerability arises from over-reliance on visual appearance rather than chart semantics. By explicitly extracting chart structure before answering, models become less susceptible to several forms of deception, particularly those involving axis manipulation and inappropriate encodings. However, the effectiveness of this strategy depends heavily on the quality of the extracted representation. Stronger models benefit substantially from the additional grounding, whereas weaker models may propagate extraction errors into downstream reasoning. These findings indicate that future mitigation strategies should focus not only on reasoning but also on improving the reliability of chart interpretation and structured representation learning.

\textbf{Structured chart representations may support human verification.} Unlike standard end-to-end VLM reasoning, the proposed framework exposes intermediate chart representations, including reconstructed data values, axis properties, and visual encodings. These representations could allow users to inspect chart structure and identify potentially misleading design choices before accepting a model’s conclusion. Such transparency may be particularly valuable in settings where deceptive charts can influence public opinion, such as social media and online news. These findings suggest that exposing intermediate chart representations may be an important complement to improving model accuracy alone.

\textbf{Limitations.}
Our work has several limitations. First, the benchmark is constructed from programmatically generated charts rather than manually collected real-world misleading visualizations. This design choice enables controlled paired comparisons, where faithful and misleading charts differ only in a single deceptive intervention. Although we use real-world data and widely adopted visualization libraries to preserve realism, naturally occurring misleading charts may contain additional contextual cues, annotations, and design artifacts that are not captured by our benchmark. Second, our mitigation approach is evaluated only as an inference-time strategy and remains dependent on the capabilities of the underlying VLM. As demonstrated by some smaller models, errors introduced during chart extraction can propagate to downstream reasoning and reduce the effectiveness of the mitigation pipeline. Finally, our benchmark focuses exclusively on design-based deception. Real-world misinformation often combines visual manipulation with textual framing, captions, and narrative context, which are outside the scope of the current benchmark.

 \section{Conclusion and Future Work}
In this work, we presented a systematic study of how misleading chart designs affect the reasoning behavior of vision-language models.  We introduced \benchmark, a controlled benchmark covering eight major categories of deceptive visualization designs, and evaluated 10 state-of-the-art open-source and proprietary VLMs across 48,000 responses. Our findings reveal that current VLMs remain consistently vulnerable to misleading visual encodings, even when the underlying data remain unchanged. To better characterize these failures, we introduced the \textit{Deception Score}, which isolates deception-induced error from baseline chart-understanding error and provides a more faithful measure of model susceptibility. Together, these results show that strong chart understanding performance does not necessarily imply robustness to deceptive visual design. Finally, we investigated whether robustness can be improved at inference time through structured reasoning. Our results demonstrate that a multi-agent framework based on explicit chart extraction and downstream reasoning reduces deception-induced error for most models, particularly stronger VLMs. These findings suggest that encouraging models to reason from chart semantics rather than visual appearance alone is a promising direction for improving robustness.

Several promising directions emerge from this work. First, fine-tuning VLMs for robustness against misleading visualizations could provide a stronger, albeit more computationally expensive, mitigation strategy than inference-time interventions alone. Second, the impact of misleading VLM responses on human interpretation should be studied more extensively, particularly in settings such as social media where deceptive charts can rapidly shape public perception. Third, future reasoning architectures could internalize the metadata-generation process, enabling models to reason more explicitly over chart structure and visual design choices rather than relying primarily on visual appearance. 
By introducing a benchmark, metric, and mitigation framework for chart deception, we hope to establish robustness to misleading visualizations as an important evaluation criterion for chart-understanding systems. Ultimately, trustworthy chart understanding requires not only answering questions correctly, but also remaining robust when visual design choices distort the story the data appear to tell.

\bibliographystyle{IEEEtran}
\bibliography{template2}



\begin{IEEEbiography}[{\includegraphics[width=1in,height=1.25in,clip,keepaspectratio]{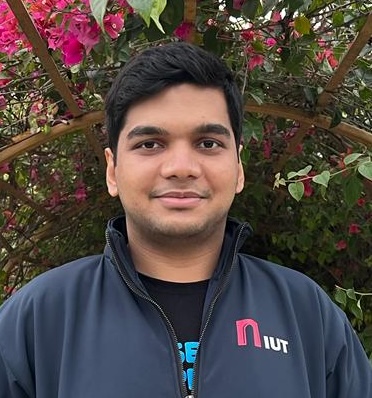}}]{Ridwan Mahbub} is an MSc student at York University and a Graduate Research Assistant at the Intelligent Visualization Lab. His research focuses on evaluating and improving large language models and vision-language models for data visualization tasks, including chart understanding, data video generation, misleading visualization analysis, and Agentic AI for visualization. He develops benchmarks, evaluation frameworks, and mitigation approaches to study and improve model robustness in visualization-centric settings. He has published papers in reputed venues, including ACL, EMNLP, EACL, and IEEE VIS, and received the IEEE VIS 2025 Best Short Paper Award. Ridwan has also served as a peer reviewer for venues such as ACL and EMNLP.

\end{IEEEbiography}

\begin{IEEEbiography}[{\includegraphics[width=1in,height=1.25in,clip,keepaspectratio]{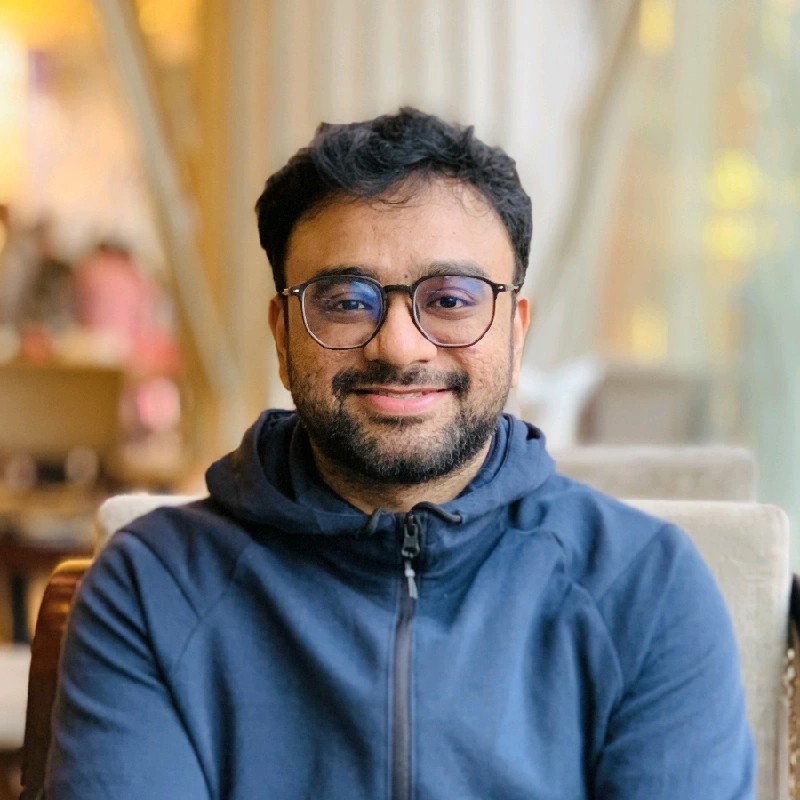}}]{Mohammed Saidul Islam} is an Associate Applied Machine Learning Specialist at the Vector Institute and previously worked as a Graduate Research Assistant at the Intelligent Visualization Lab, York University. He obtained his master’s degree in Computer Science from York University and his bachelor’s degree in Computer Science and Engineering from the Islamic University of Technology. His research focuses on reliable evaluation of large language models and vision-language models, multimodal reasoning, chart understanding, data-driven storytelling, text-to-visualization, and Agentic AI. He has published papers in reputed conference proceedings, including ACL, EMNLP, EACL, and LREC-COLING. He also received the IEEE VIS Best Short Paper Award. In addition to his research, Saidul has served as a peer reviewer for ACL, EMNLP, NAACL, and ACL ARR.

\end{IEEEbiography}

\begin{IEEEbiography}[{\includegraphics[width=1in,height=1.25in,clip,keepaspectratio]{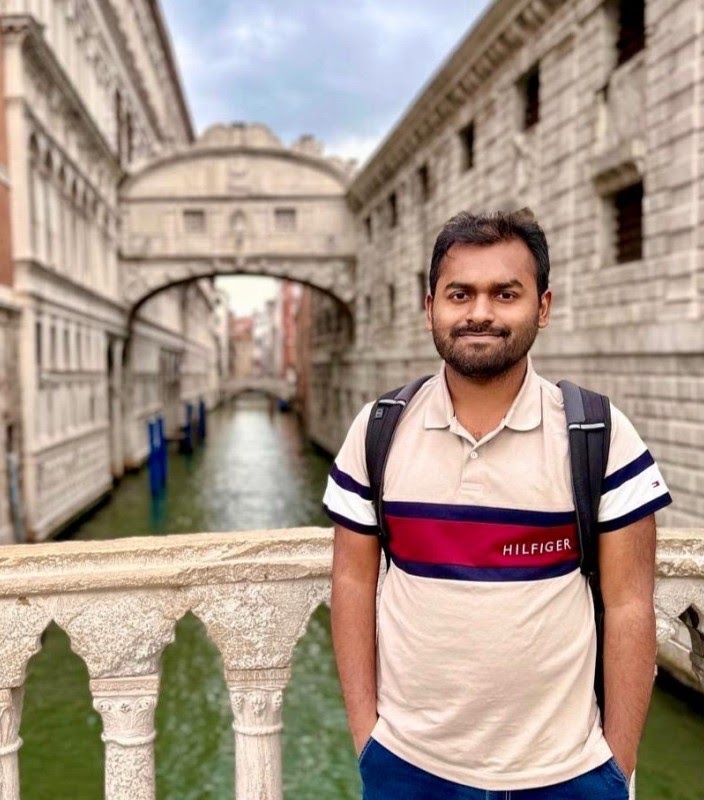}}]{Md Tahmid Rahman Laskar} has worked as a Senior Applied Scientist at Dialpad and currently also pursuing his doctoral studies in computer science at York University. Prior to that, he obtained his master’s degree in Computer Science from York University. He has published research papers in several reputed journals, such as Computational Linguistics, ACM Transactions on Cyber–Physical Systems, ACM Computing Surveys, Computers in Human Behavior, Computers in Biology and Medicine, and the Journal of Ambient Intelligence and Humanized Computing. He has also published more than 40 research papers in referred conference proceedings, which include ACL, ICML, SIGIR, EMNLP, NAACL, COLING, EACL, IEEE VIS etc. He was also awarded the best paper award at the DaSH Workshop in EMNLP 2022 and in IEEE VIS 2025. In addition, he was a co-organizer of the BLP Workshop at AACL-IJCNLP 2025, currently serving as an Area Chair at ACL Rolling Review and received the outstanding area chair award at EACL 2026. Tahmid's current research is focused on ensuring reliable evaluation of Large Language Models.

\end{IEEEbiography}

\begin{IEEEbiography}[{\includegraphics[width=1in,height=1.25in,clip,keepaspectratio]{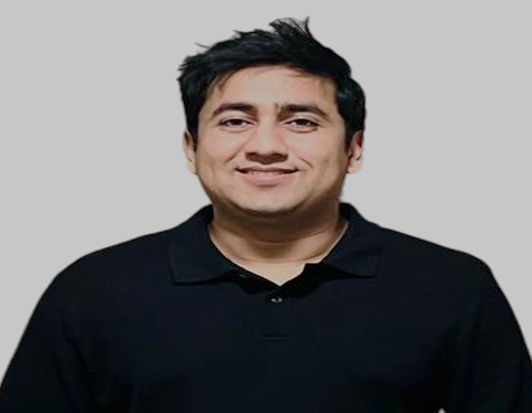}}]{Mizanur Rahman}
is a graduate student at York University. His research interests include large language models, vision-language models, multimodal reasoning, chart understanding, and data visualization. He has published papers in reputed journals and conferences, including ACM Computing Surveys, Journal of Applied Physics, ACL, EMNLP, EACL, and IEEE VIS. He also received the IEEE VIS Best Short Paper Award. In addition, he has served as a reviewer for ACM Computing Surveys, ACL, EMNLP, and EACL.
\end{IEEEbiography}

\begin{IEEEbiography}[{\includegraphics[width=1in,height=1.05in,clip,keepaspectratio]{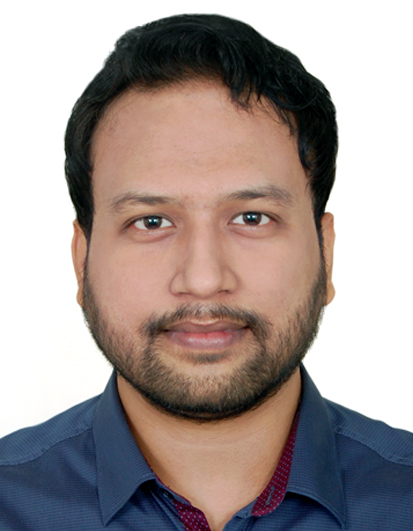}}]{Mir Tafseer Nayeem} is a PhD Candidate in Computing Science at the University of Alberta, Canada. His research centers on language modeling, with a particular focus on user-centric aspects of large language models. He studies how user attributes, such as age, language, dialect, intent, and preferences, can be integrated into the development of LLMs. His work has been published in leading journals and conferences, including ACL, COLM, EMNLP, AAAI, NAACL, EACL, CIKM, and IEEE VIS etc. His research has also been recognized with multiple paper awards at top conferences, including EMNLP, COLING, and IEEE VIS. He is currently serving as an Area Chair for ACL-family conferences, including ACL, EMNLP, NAACL, and EACL, through ACL Rolling Review.

\end{IEEEbiography}

\begin{IEEEbiography}[{\includegraphics[width=1in,height=1.25in,clip,keepaspectratio]{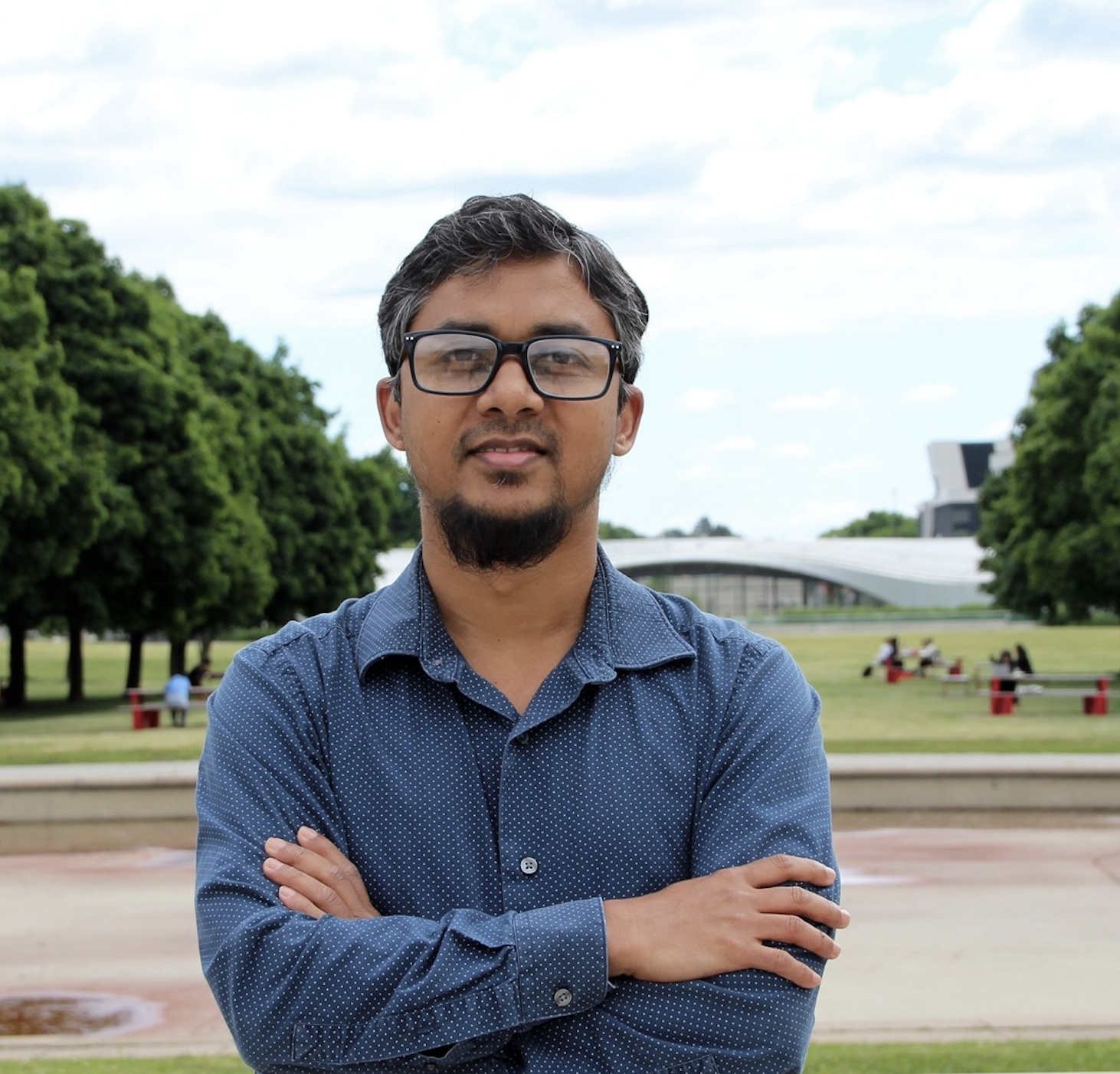}}]{Enamul Hoque} (Senior Member, IEEE) is an Associate Professor in the School of Information Technology at York University, Canada.
Before joining York University, he was a postdoctoral fellow in Computer Science at Stanford University. He received the Ph.D. degree in Computer Science from the University of British Columbia. 
His research bridges information visualization, natural language processing, and human-computer interaction to make data analytics more accessible and inclusive.  He has received distinctions including the Dean’s Award for Research Excellence, as well as Best Paper and Honorable Mention awards. Dr. Hoque has published nearly 100 papers in top venues such as IEEE VIS, ACL, EMNLP, and CHI, and has delivered numerous tutorials, panel discussions, and invited talks on vision-language models for data visualization. He serves as an Area Chair for the ACL Rolling Review and as a program committee member for IEEE VIS. His research has been supported by NSERC, the Canada Foundation for Innovation, and the National Research Council Canada. 

\end{IEEEbiography}




\vfill

\end{document}